\documentclass[a4paper]{spie}  

\usepackage{amsmath,amsfonts,amssymb}
\usepackage{graphicx}
\usepackage{hyperref}
\usepackage{float}
\usepackage{subfigure}
\usepackage{multirow}

\title{See Clearer at Night: Towards Robust Nighttime Semantic Segmentation through Day-Night Image Conversion}
\author{Lei Sun, Kaiwei Wang, Kailun Yang, and Kaite Xiang}
\affil{State Key Laboratory of Modern Optical Instrumentation, Zhejiang University, China}
\authorinfo{Correspondence: wangkaiwei@zju.edu.cn}
\begin{document} 
\maketitle

\begin{abstract}
In recent years, intelligent driving navigation and security monitoring have made considerable progress with the help of deep Convolutional Neural Networks (CNNs). As one of the state-of-the-art perception approaches, semantic segmentation unifies distinct detection tasks widely desired by both autonomous driving and security monitoring. Currently, semantic segmentation shows remarkable efficiency and reliability in standard scenarios such as daytime scenes with favorable illumination conditions. However, in face of adverse conditions such as the nighttime, semantic segmentation loses its accuracy significantly. One of the main causes of the problem is the lack of sufficient annotated segmentation datasets of nighttime scenes. In this paper, we propose a framework to alleviate the accuracy decline when semantic segmentation is taken to adverse conditions by using Generative Adversarial Networks (GANs). To bridge the daytime and nighttime image domains, we made key observation that compared to datasets in adverse conditions, there are considerable amount of segmentation datasets in standard conditions such as BDD and our collected ZJU datasets. Our GAN-based nighttime semantic segmentation framework includes two methods. In the first method, GANs were used to translate nighttime images to the daytime, thus semantic segmentation can be performed using robust models already trained on daytime datasets. In another method, we use GANs to translate different ratio of daytime images in the dataset to the nighttime but still with their labels. In this sense, synthetic nighttime segmentation datasets can be generated to yield models prepared to operate at nighttime conditions robustly. In our experiment, the later method significantly boosts the performance at the nighttime evidenced by quantitative results using Intersection over Union (IoU) and Pixel Accuracy (Acc). We show that the performance varies with respect to the proportion of synthetic nighttime images in the dataset, where the sweet spot corresponds to most robust performance across the day and night. The proposed framework not only makes contribution to the optimization of visual perception in intelligent vehicles, but also can be applied to diverse navigational assistance systems.
\end{abstract}
\section{INTRODUCTION}
\label{sec:intro}

\begin{figure}[ht]
   \centering
   \includegraphics[width = 0.98\textwidth]{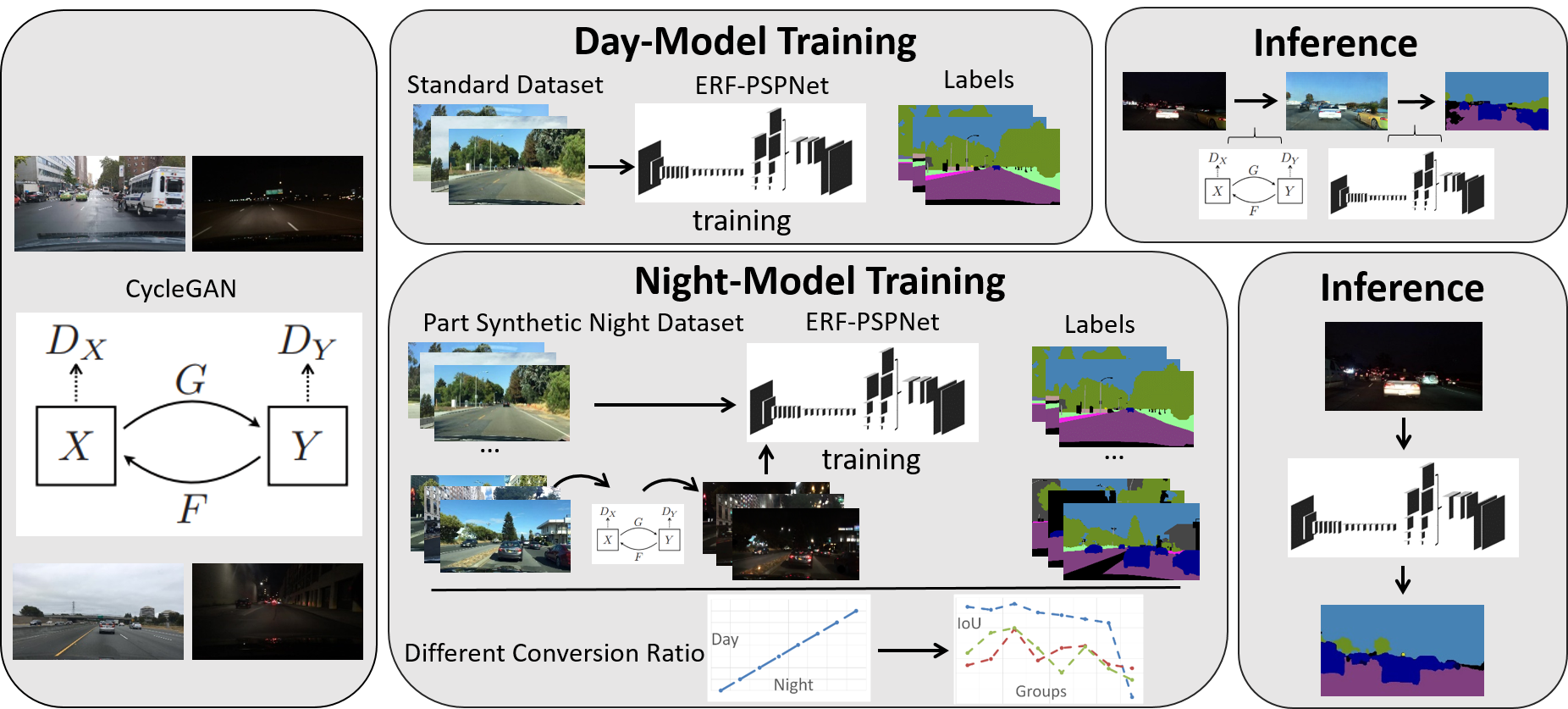}
   \caption{\label{fig:main_frame}
   The figure shows the main frame of our work. On the left, a day-night converter is trained using unpaired datytime and nighttime images. The first row shows our first method: training a day-model and converting night domain images to the day domain before inference. Bottom row shows our second method: converting different ratios of images in the training set to nighttime images to train a night model. The ratio of synthetic nighttime images determines the model's accuracy in testing sets.}
\end{figure}

Vision tasks like object detection and semantic segmentation (i.e. pixel-wise scene classification) are always the key points in security monitoring and autonomous driving. Convolutional Neural Networks(CNNs) have fueled the development of these methods thanks to the availability of larger datasets and computationally-powerful machines that have emerged in recent years. Semantic segmentation, which unifies distinct detection tasks with a single consumer camera\cite{yang2018unifying,yang2018unifyingterrainawareness,yang2018intersection} makes RADAR and LiDAR sensors become the second choices, freeing scene perception from complex multi-sensor fusion\cite{long2018fusion,pfeuffer2019robust,yang2016expanding}. Some state-of-the-art CNN methods like PSPNet\cite{zhao2017pyramid}, RefineNet\cite{lin2017refinenet}, DeepLab\cite{chen2017deeplab}, and ACNet\cite{hu2019acnet} perform semantic segmentation with very high accuracies. In order to apply semantic segmentation to autonomous driving and security monitoring, we have proposed ERF-PSPNet\cite{yang2018unifying,yang2019can}, a high-accuracy real-time semantic segmentation method in previous work, which is more computationally efficient than most of the state-of-the-art methods.

All these perception algorithms are designed to operate on images taken at daytime under good illumination conditions\cite{yang2019robustifying,cordts2016cityscapes,neuhold2017mapillary}. However, outdoor applications can hardly escape from challenging weather and illumination conditions. One of the reasons why computer vision system based on semantic segmentation have not been widely applied yet is because that it can not deal with adverse conditions. For example, semantic segmentation using visible light camera performs unsatisfactorily in the nighttime for the reason that when under extremely weak illuminance, the structure, texture and color features of objects change drastically. These features can either disappear because of the lack of the illuminance or being highly disturbed by artificial light. Thus, how to enhance the robustness of semantic segmentation has been an important issue in the computer vision domain\cite{romera2019bridging}. In this work, we focus on improving nighttime semantic segmentation performance.

There are some researchers that have proposed to use Far-Infrared (FIR) camera instead of visible light camera\cite{li2019segmenting}. FIR camera is a feasible measure, but they are expansive and only provide low-resolution images. In addition, there are rare FIR semantic segmentation datasets. In the other way, visible light camera is much cheaper, and there are a large number of datasets of daytime images. For these reasons, we choose visible light camera as the image acquisition device. On the other hand, high-accuracy semantic segmentation models are trained from large scale of annotated images. But annotating nighttime images requires extensive time and human effort, and it is impossible to annotate images at the pixel wise for all the other adverse conditions.

\begin{figure}[ht]
   \centering
   \includegraphics[width = 0.5\textwidth]{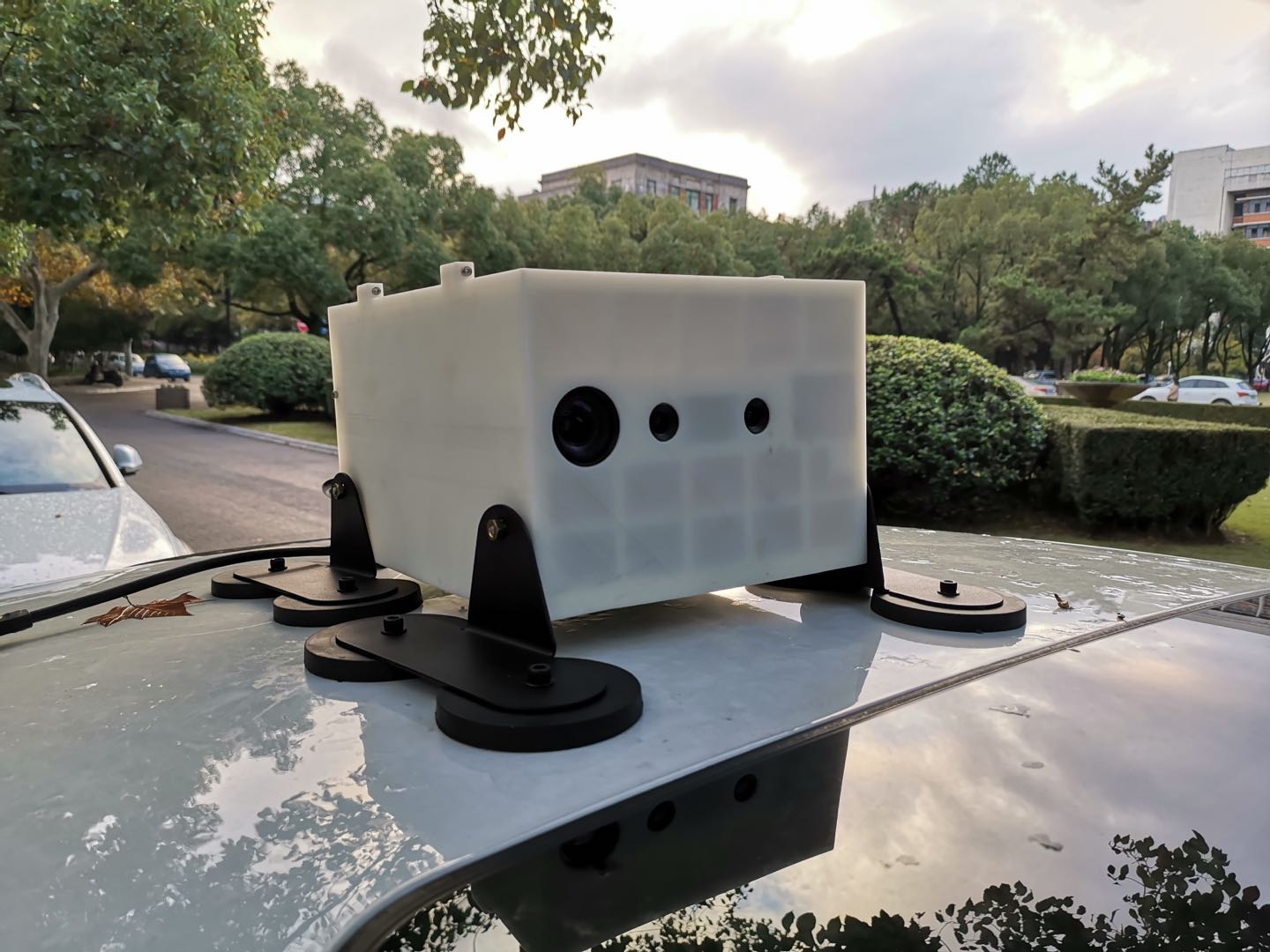}
   \caption{\label{fig:sensor}
   Our Multi-Modal Stereo Vision Sensor on the top of an instrumented vehicle used to capture the ZJU dataset.}
\end{figure}

In this paper, we mainly propose a main frame (see Fig.~\ref{fig:main_frame}) to overcome the problem of large accuracy downgrade from daytime to nighttime in semantic segmentation. Inspired by the idea of Generative Adversarial Networks (GANs)\cite{goodfellow2014generative}, nighttime images are converted on-the-fly during inference to the daytime domain as a pre-processing step of the first proposed method. In the other way, we augment an original large-scale semantic segmentation dataset such as the BDD database\cite{yu2018bdd100k} by translating part of the daytime images to nighttime images. Among the experiments, the feasibility of improving robustness of semantic segmentation is validated. In addition, we record a dataset in the Yuquan Campus of Zhejiang university with both day and night images as well as GPS information by using our Multi-Modal Stereo Vision Sensor\cite{sun2019multi} (see Fig.\ref{fig:sensor}), which has been made publicly available.

\section{RELATED WORKS} 

\subsection{Semantic Understanding of the Road Scene}
\label{sec:title}
Semantic segmentation is important in understanding the content of images and finding target objects, and this technique is vital for the field of automatic driving\cite{xiang2019importance,xiang2019comparative}. Currently, most of state-of-the-art semantic segmentation models are based on fully convolutional end-to-end networks\cite{shelhamer2016fully}. Inspired by SegNet\cite{badrinarayanan2017segnet}, semantic segmentation models usually follow an encoder-decoder network architecture. The encoder is a vanilla CNN which is trained to classify the input, and the decoder is used to upsample the output of the encoder to the same size as the input image\cite{zhao2017pyramid,chen2017deeplab,chen2014semantic,chen2017rethinking,chen2018encoder}. Further, more efficient networks were proposed to achieve the goal of real-time semantic segmentation\cite{romera2019bridging,paszke2016enet,romera2017erfnet}. Our works are based on ERF-PSPNet\cite{yang2018unifying,yang2019can}, a state-of-the-art semantic segmentation network designed for navigation assistance systems.

\subsection{Model Adaption}
Generally, CNNs only learn features from the domain of training datasets, and may perform much worse in a different domain. This is also the reason why semantic segmentation model trained in daytime domain drops accuracy in the nighttime domain. To improve the generalization of convolutional neural network, many methods were proposed. Most commonly, data augmentation techniques like random cropping, random rotation and flipping are used to make networks perform stably in unfamiliar domains\cite{yang2019robustifying}. Effective use of synthetic data has been preliminarily studied in\cite{sadat2018effective,xu2019semantic}. There is another domain adaptation-based approach that was proposed to adapt semantic segmentation models from synthetic images to real environments\cite{sankaranarayanan2018learning}. The other attempts that are most close to our work, which also improve the model robustness at the nighttime, make use of twilight images to transfer knowledge from standard daytime conditions to nighttime images\cite{dai2018dark,sakaridis2019semantic}. Similar efforts were also made to address robust foggy scene parsing\cite{sakaridis2018model,dai2019curriculum} and rainy scene semantic segmentation\cite{porav2019can,hu2019depth}. Unsupervised learning has also been frequently leveraged to pre-process input images, in order to prevent performance from degrading catastrophically when the input domain differs significantly from previously seen domains\cite{romera2019bridging,porav2019don}. Specifically, this research line is also highly related to topological localization\cite{porav2019don,larsson2019cross}, where modern visual localizers like\cite{lin2018visual,cheng2019panoramic} can also benefit from the input adaptation to perform more reliably against variation challenges. More recently, model distillation/imitation were applied to make model behave stable in unseen domains\cite{hinton2015distilling,gupta2016cross}.

\subsection{Image Stylization}
Since I.~Goodfellow et al. proposed Generative Adversarial Networks (GANs)\cite{goodfellow2014generative}, GANs have become the most promising method for image stylization. Formally, GANs simultaneously contain two models: a generative model G that captures the critical distribution, and a discriminative model D that estimates the probability that a sample came from the training data rather than generated by G. Although state-of-the-art GANs like Pix2Pix\cite{isola2017image} perform impressively for style transfer, the training data in both domains have to be pre-formatted into a single X/Y image pair that holds tight pixel-wise correlation. A recently proposed CycleGAN\cite{zhu2017unpaired} is designed to perform a full translation cycle, and make it possible to make use of images in two different domains without paring images, which is suitable for our work to translate image across daytime and nighttime domains.

\section{METHOD}
\label{sec:sections}

In our work, two methods are proposed to narrow the gap between daytime and nighttime images in semantic segmentation. These methods respectively correspond to converting nighttime images to daytime images and the vice versa. Fig.~\ref{fig:main_frame} shows our framework. In both methods, we train a CycleGAN to perform domain converting. In the first method, we convert nighttime images to daytime, in order to shift the images to the suitable domain. Next, the ERF-PSPNet\cite{yang2018unifying,yang2019can} trained on daytime images predict semantic maps in the inference. In the second method, CycleGAN converts parts of daytime images in the training set to nighttime images to extend the domain coverage of the datasets. After that, the adapted training dataset with a certain percent of nighttime images is used to train ERF-PSPNet, in order to improve its performance at the nighttime.
 
\subsection{Training CycleGAN for Night-day Domain Converting}

In this subsection, our work is training a GAN to translate nighttime images to daytime or reverse. Image-to-image translation is a class of vision problems in which the goal is to learn the mapping between input images and output images by using a training set of aligned image pairs. But for our task, large scale of paired training data is not available. Although collecting paired datasets by ourselves is theoretically possible, it is impractical to collect datasets for every different styles of scenes. What we need is an universal night-day domain converter that can be utilized at the dataset level.

CycleGAN\cite{zhu2017unpaired} is an approach for learning to translate an image from a source domain to a target domain in the absence of paired examples, which suits our needs. CycleGAN contains two sets of GANs. Each GAN contains a generator and a discriminator. Generator and discriminator make translator, to translate image from domain X to domain Y or vice versa. Two GANs represent two generators: F and G, and they are inverses of each other. we have trained both the mapping G and F simultaneously and adding a cycle consistency loss that encourages \begin{equation}F(G(x))\approx x\end{equation} and \begin{equation}G(F(y))\approx y\end{equation}
This loss makes the unpaired image-to-image translation possible. 

In our work, we select 6000 daytime images and 6000 nighttime images from the BDD100K dataset\cite{yu2018bdd100k}, as two image domains to train CycleGAN. Limited to GPU memory, we resize images to 480$\times$270 to train our CycleGAN. In this way, we obtain our day-night and night-day converters.

\subsection{Converting Images to the Daytime During Inference}

The first option is to convert nighttime images to daytime images. More specifically, nighttime images acquired by the camera are converted to synthetic daytime images, which is the suitable domain for semantic segmentation. 

This method does not need to train the semantic segmentation model again. In other words, the advantage of the method is that we can make use of the original weights in the trained ERF-PSPNet, which has already been demonstrated to be stable in most datasets\cite{xiang2019importance} and actual scenarios\cite{yang2019can,yang2019robustifying}. Additionally, the night-day conversion and semantic segmentation inference are separated, which makes it easier to adjust.

But the computational cost of the inference process is increased, which is a disadvantage for real-time semantic segmentation. For each inference process, the converted images from CycleGAN are fed into ERF-PSPNet. While efficient segmentation network is readily available for predicting accurate semantic maps, the forward pass of CycleGAN costs nearly 1 second for each 480$\times$270 image. This is too slow for a real-time semantic segmentation model like ERF-PSPNet and the system losses its real-time performance. Another disadvantage is that compared to original images, the synthetic image produced by GAN may be biased. For example, GAN may convert far-away buildings to trees etc.

\subsection{Generating Nighttime Images to Expand the Training Set}

The second option is to convert daytime images in the BDD10K training set with segmentation labels to nighttime images. Then, the training set with part of synthetic nighttime images is fed into ERF-PSPNet, with focal loss as the loss function\cite{lin2017focal,yang2018predicting}. The idea comes from the lack of the nighttime datasets with precise segmentation labels. 

The advantage of the method is that for a trained model, no extra calculation is introduced in the inference process. For this reason, ERF-PSPNet can keep its real-time property. In our experiment, we conduct experiment to explore how the ratio of synthetic nighttime images influences the accuracy of the semantic segmentation model. Based on the experiment, we perform the discussion on the necessity of adding nighttime images with precise labels in the training dataset.

However, the disadvantage of the method is the time-consuming process of re-training model, and the model may not always be robust for all kinds of environments. In addition, because of the large scale of parameters in CycleGAN, we have to resize images in BDD100K to 480$\times$270 to train the GAN. In this way, GAN can only produce images with the size of 480$\times$270, which is below 1280$\times$720, the resolution of images in BDD100K. So we have to upsample the synthetic images to 1280$\times$720 before feeding into the segmentation model. Such operation makes unavoidable influence on the accuracy of the final prediction result.

\section{EXPERIMENTS}

\subsection{Datasets and Training}

The first step of our experiment is to train a GAN converting nighttime images to daytime images and vice versa. As describe above, paired images are not necessary for CycleGAN\cite{zhu2017unpaired}, but the datasets must contain diverse driving images at both daytime and nighttime. As nighttime images are absent in most mainstream datasets for semantic segmentation like Cityscapes\cite{cordts2016cityscapes} and Mapillary Vistas\cite{neuhold2017mapillary}, finally we utilize the lately released BDD datasets\cite{yu2018bdd100k}. 

BDD datasets contain two parts: BDD100K and BDD10K. The former contains 10,0000 driving images captured in diverse conditions and different time with GPS information, object detection annotation, lane and drivable area annotation. The latter one contains 9,000 pixel-wise semantically annotated images and 1,000 test images with 19 labeled classes.

To qualitatively verify our methods for real-world applications, we also collect daytime and nighttime images in the Yuquan Campus of Zhejiang University by using our Multi-Modal Stereo Vision Sensor\cite{sun2019multi} fixed on the top of an instrumented car. More than 1,500 images were acquired for our research. Furthermore, we utilize Nighttime Driving Test dataset provided by D.~Dai et al.\cite{dai2018dark}, which contains 50 nighttime images with precise segmentation annotation. Table.~\ref{tab:datasets} and Fig.~\ref{fig:datasets} show details about these datasets and all three datasets are again itemized below:

\begin{figure}[ht]
   \centering
   \subfigure[BDD dataset]{
   \begin{minipage}[b]{0.32\linewidth}
   \includegraphics[width=1\linewidth]{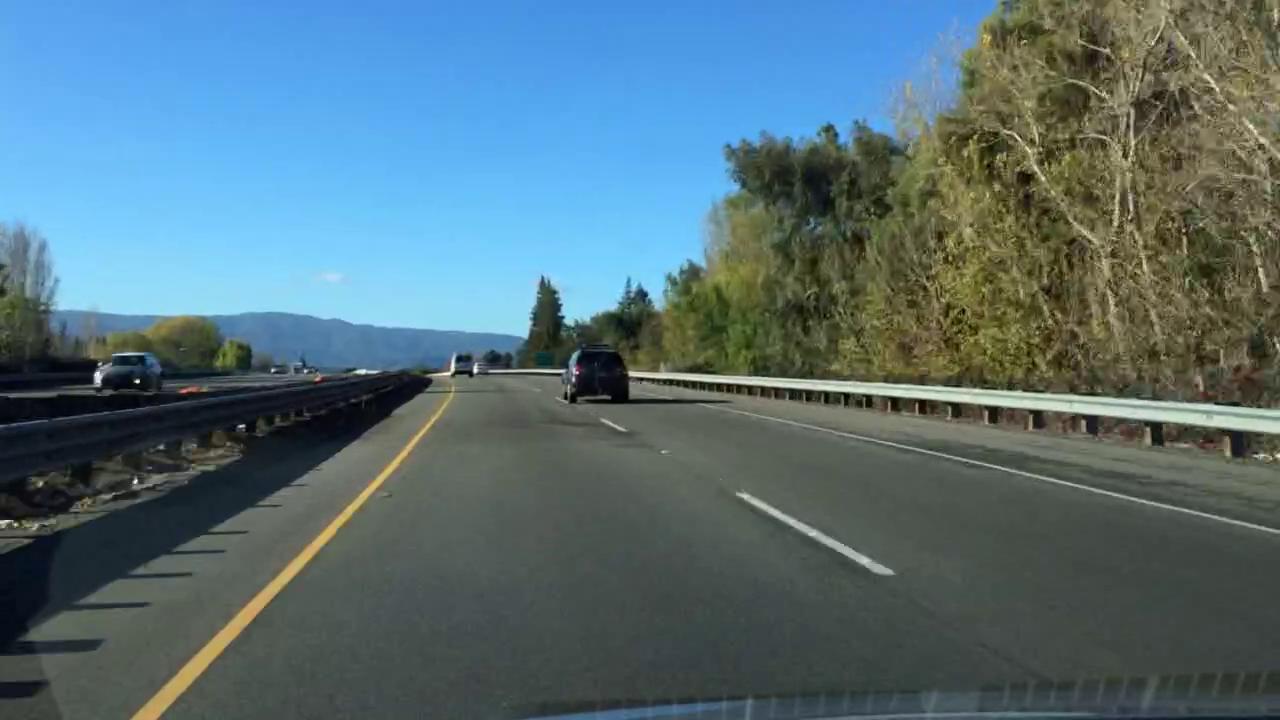}\vspace{0pt}  
   \includegraphics[width=1\linewidth]{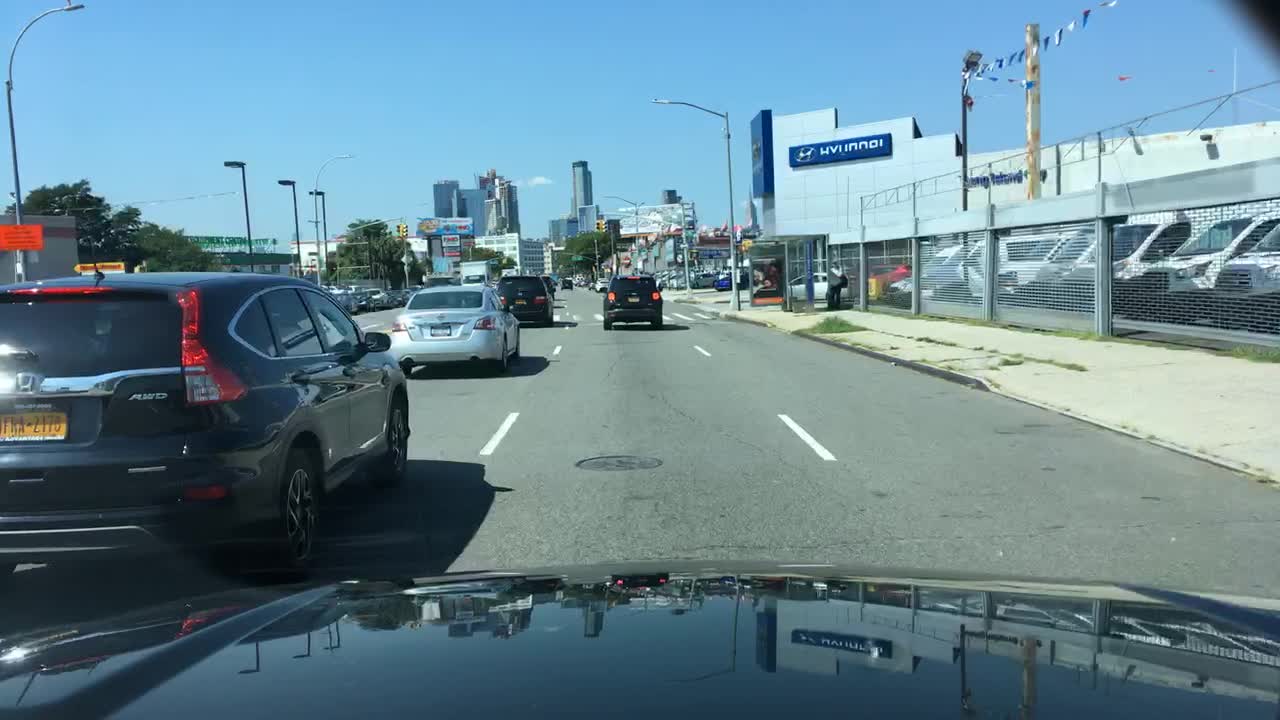}
   \end{minipage}}\hspace{-1pt}  
   \subfigure[Nighttime Driving dataset]{
   \begin{minipage}[b]{0.32\linewidth}
   \includegraphics[width=1\linewidth]{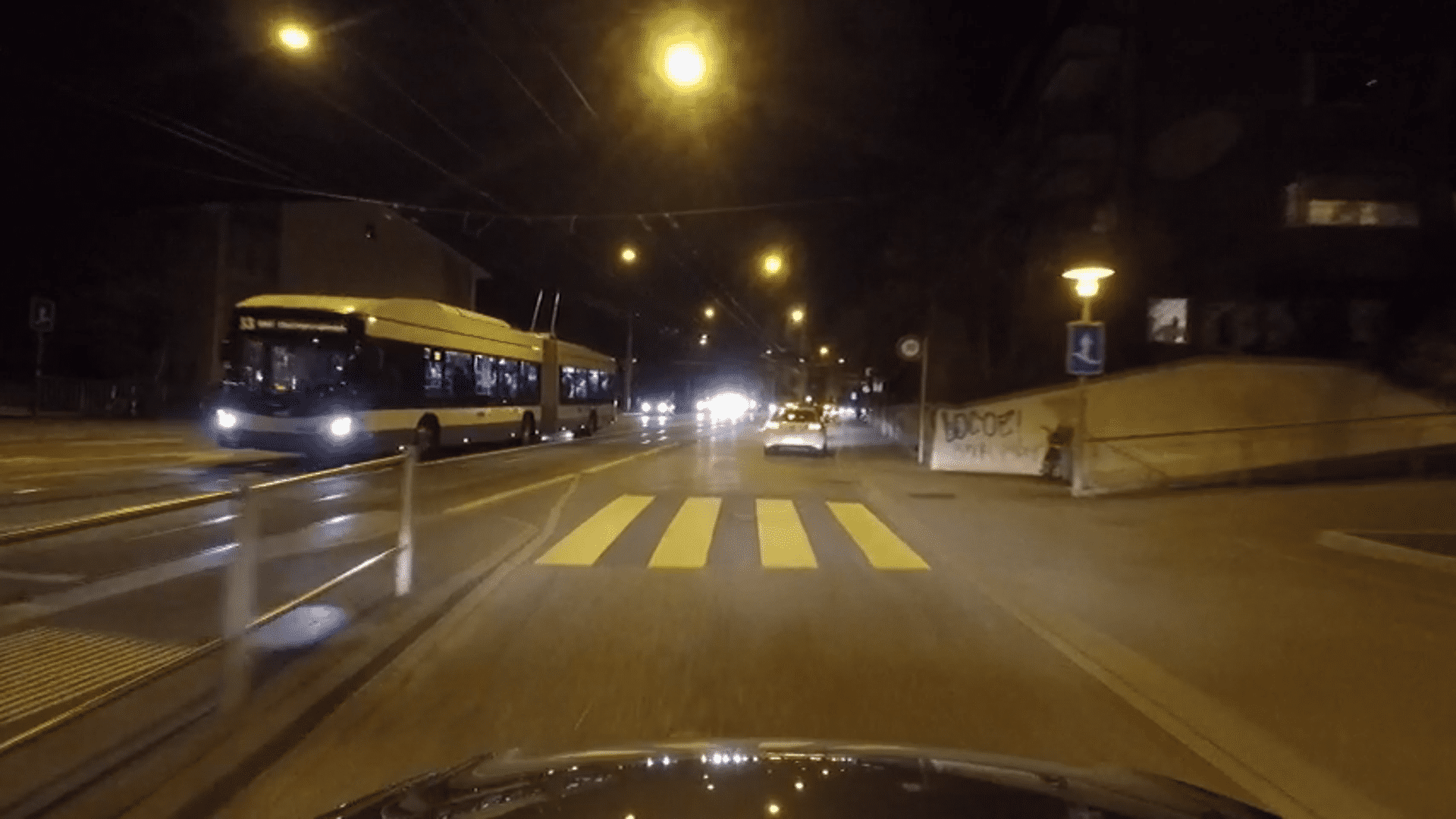}\vspace{0pt}
   \includegraphics[width=1\linewidth]{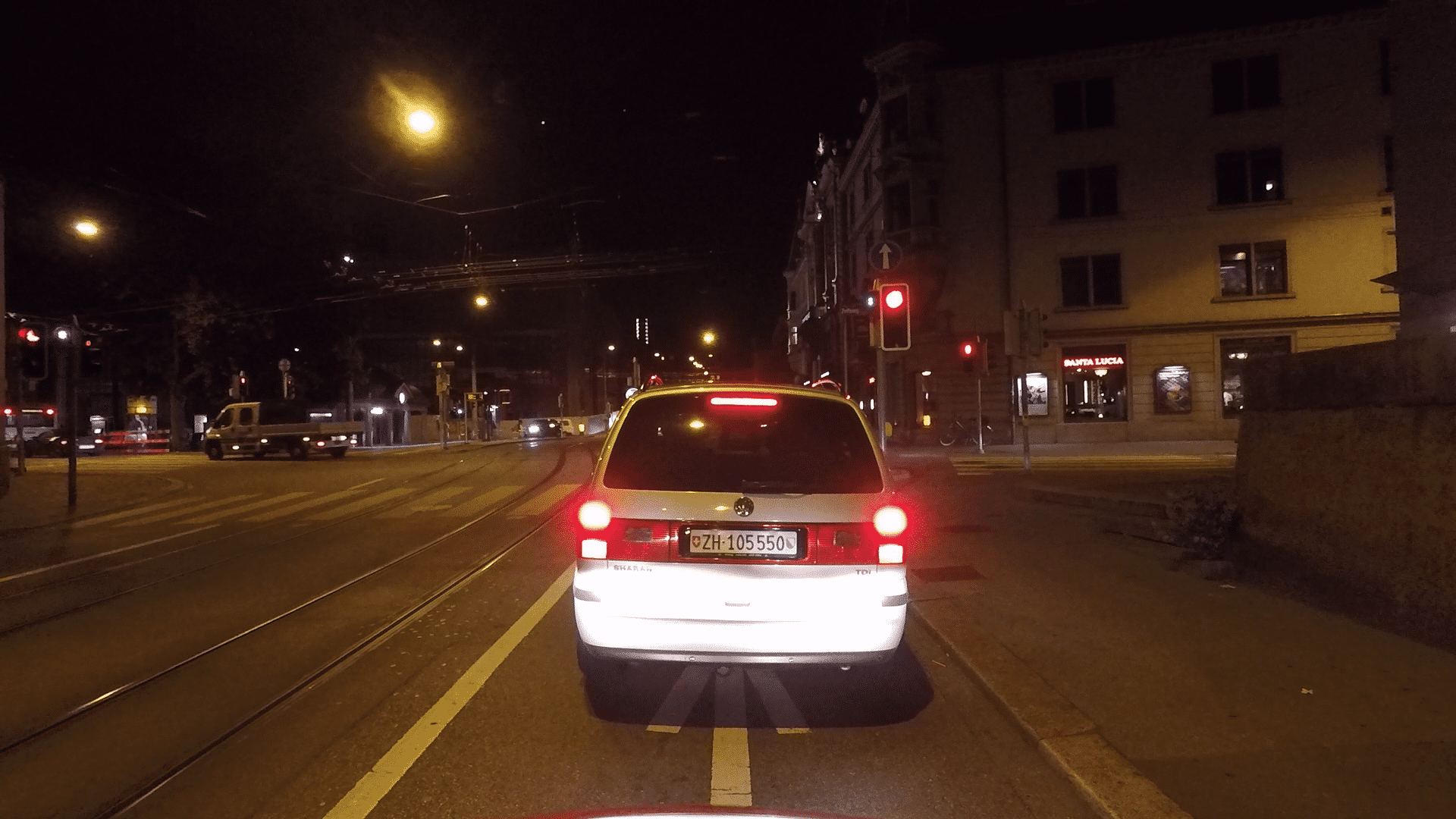}
   \end{minipage}}\hspace{-1pt}
   \subfigure[ZJU dataset]{
   \begin{minipage}[b]{0.32\linewidth}
   \includegraphics[width=1\linewidth]{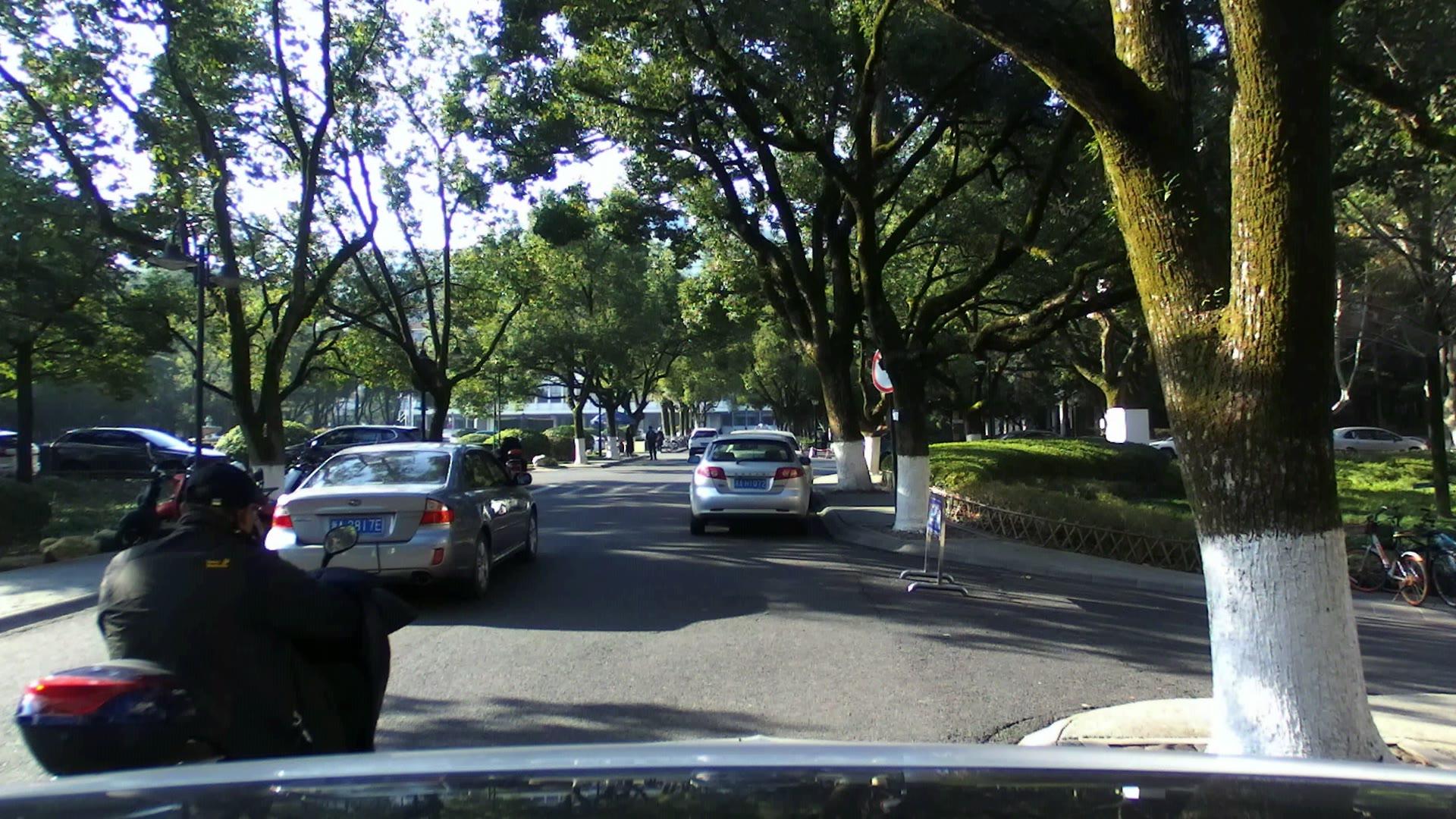}\vspace{0pt}
   \includegraphics[width=1\linewidth]{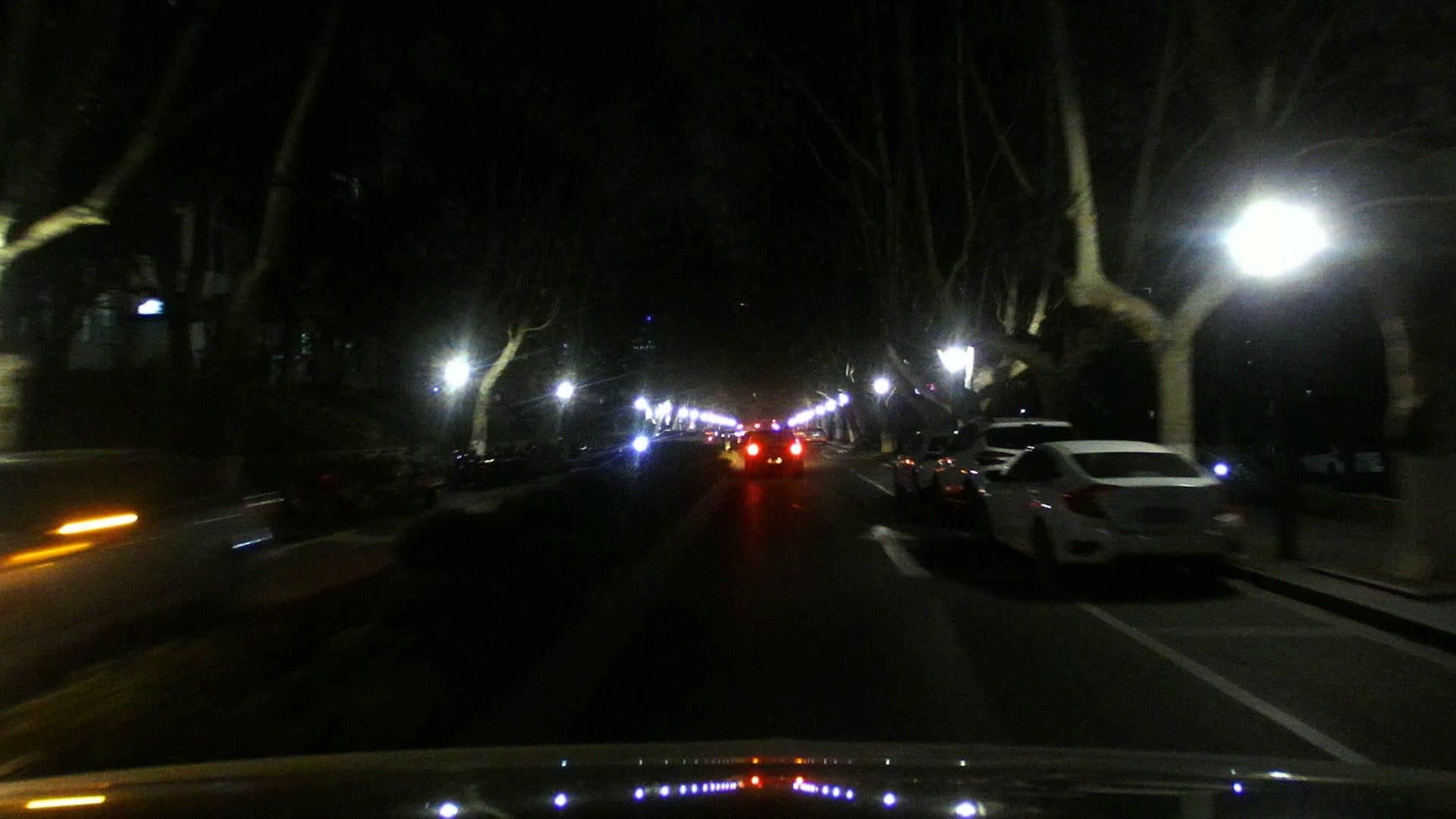}
   \end{minipage}}\hspace{-1pt}
   \caption[datasets]
   { \label{fig:datasets}
   Examples from three datasets. The streetscape of images in ZJU dataset varies from that in BDD dataset and Nighttime Driving dataset.}
\end{figure}

\begin{description}
   \item[BDD Dataset\cite{yu2018bdd100k}.] In total, BDD dataset have 100,000 driving images collected from more than 50,000 rides, covering New York, San Francisco Bay Area, and Berkeley. The dataset contains diverse scene scenarios such as city streets, residential areas, and highways. BDD100K contains plenty of nighttime images, making it possible to train a day-night converter. However BDD10K contains only 32 nighttime images with pixel-wise labels.
   \item[ZJU Dataset.] The dataset was captured in Zhejiang University, Yuquan Campus (Hangzhou, China) with our Multi-Modal Stereo Vision Sensor\cite{sun2019multi}. During the nighttime collection, we capture the images under two settings: with headlight-illumination and without headlight-illumination. The most significant feature of this dataset is that it has more trees and pedestrians than the others, and the road is very narrow. ZJU Dataset has also been used in our preliminary work\cite{romera2019bridging} that only uses the headlight-illuminated nighttime image sequence. In this work, we use the image sequence without headlight-illumination which is more challenging. Both image sequences including daytime, nighttime with/without headlight-illumination of the ZJU dataset have been made publicly available at \url{https://github.com/elnino9ykl/ZJU-Dataset}.
   \item[Nighttime Driving Dataset\cite{dai2018dark}.] The dataset was collected during 5 rides with a car inside multiple Swiss cities and their suburbs using a GoPro Hero 5 camera, consisting of images of real driving scenes at nighttime and twilight time, with 35,000 unlabeled and 50 densely annotated images. In this paper, we only utilize the 50 annotated nighttime images for evaluation as well as comparison by taking the method proposed by D.~Dai et al.\cite{dai2018dark} as a baseline. In general, the streetscape in this dataset is very similar to BDD dataset.
 \end{description}

\begin{table}[ht]
   \caption{Main information of the three datasets.} 
   \label{tab:datasets}
   \begin{center}       
   \begin{tabular}{|l|l|l|l|l|}
   \hline
   \rule[-1ex]{0pt}{3.5ex} \multirow{2}*{\textbf{Dataset}} & \multirow{2}*{\textbf{Resolution}} & \multicolumn{2}{|l|}{\textbf{Number of Images}} & \multirow{2}*{\textbf{Comment}} \\
   \cline{3-4}  
   ~ & ~ & \textbf{Day} & \textbf{Night} & ~ \\
   \hline
   \hline
   \rule[-1ex]{0pt}{3.5ex}  BDD100K & 1280$\times$720 & 52511 & 39986 & No semantic segmentation labels \\
   \hline
   \rule[-1ex]{0pt}{3.5ex}  BDD10K & 1280$\times$720 & 7691 & 309 & Precise semantic segmentation labels \\
   \hline
   \rule[-1ex]{0pt}{3.5ex}  ZJU &  1920$\times$1080 & 1700 & 1700 & Different streetscape from other datasets \\
   \hline
   \rule[-1ex]{0pt}{3.5ex}  Nighttime Driving test &  1920$\times$1080 & 0 & 50 & Good illumination by street lamp \\
   \hline
   \end{tabular}
   \end{center}
\end{table}

Because of the huge style differences between the BDD datasets and ZJU datasets, we trained the GANs respectively for two datsets. In BDD100K, we select 6,000 daytime and 6,000 nighttime images, both in clear weather. These images are fed into CycleGAN to train a day-night converter named BDD-GAN. Because of the massive computation cost of CycleGAN, images are resized to 480$\times$270. Similarly, 850 day-time image pairs in ZJU datasets are used to train a ZJU-GAN.

Like most semantic segmentation models, ERF-PSPNet\cite{yang2018unifying,yang2019can} is composed of two parts: encoder and decoder. The encoder part has been trained on ImageNet\cite{russakovsky2015imagenet} already, and all the training tasks for ERF-PSPNet lie in the training of the decoder part of the model. In the first method, ERF-PSPNet is trained on BDD10K. Nighttime images during inference are converted on-the-fly to daytime domain by CycleGAN. In the second method, different ratios of images in training set of BDD10K are used to train ERF-PSPNet. To quantitatively validate our method, we use the 32 nighttime images with segmentation annotation in the validation set of BDD10K and 50 nighttime images with precise segmentation annotations in the Nighttime Driving Test datasets. The style of images in Nighttime Driving Test dataset is similar to BDD10K, which makes it reasonable to apply BDD-trained semantic segmentation models on it.

\subsection{Qualitative Results}

\begin{figure}[ht]
   \centering  
   \subfigure[Bad illumination condition]{
   \label{Fig.BDD-GAN sub.1}
   \includegraphics[width=0.49\textwidth]{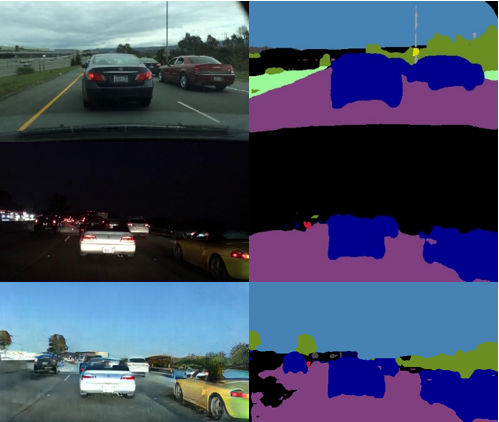}}
   \subfigure[Decent illumination condition]{
   \label{Fig.BDD-GAN sub.2}
   \includegraphics[width=0.49\textwidth]{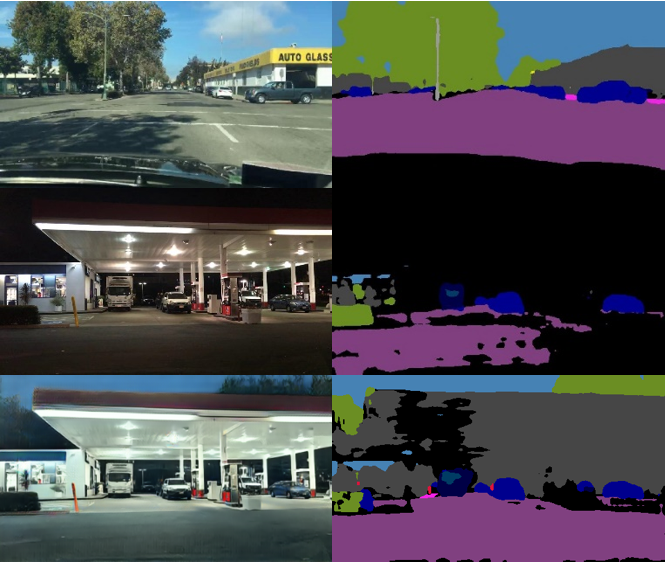}}
   
   \subfigure{
      \label{classes}
      \includegraphics[width=0.98\textwidth]{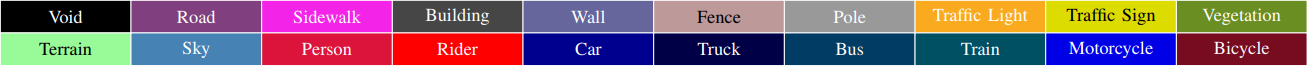}
   }
   \caption[BDD-GAN]
   { \label{fig:BDD-GAN} 
   Examples from BDD dataset (Top: day input, Mid: night input, Bottom: night-to-day converted input). Right image is better illuminated than left image. The model labeled the whole sky more precisely for synthetic daytime images than nighttime images. The night-to-day converter helps ERF-PSPNet perform better at night.}
\end{figure}

\begin{figure}[ht]
   \centering  
   \subfigure[]{
   \label{Fig.ZJU-GAN sub.1}
   \includegraphics[width=0.49\textwidth]{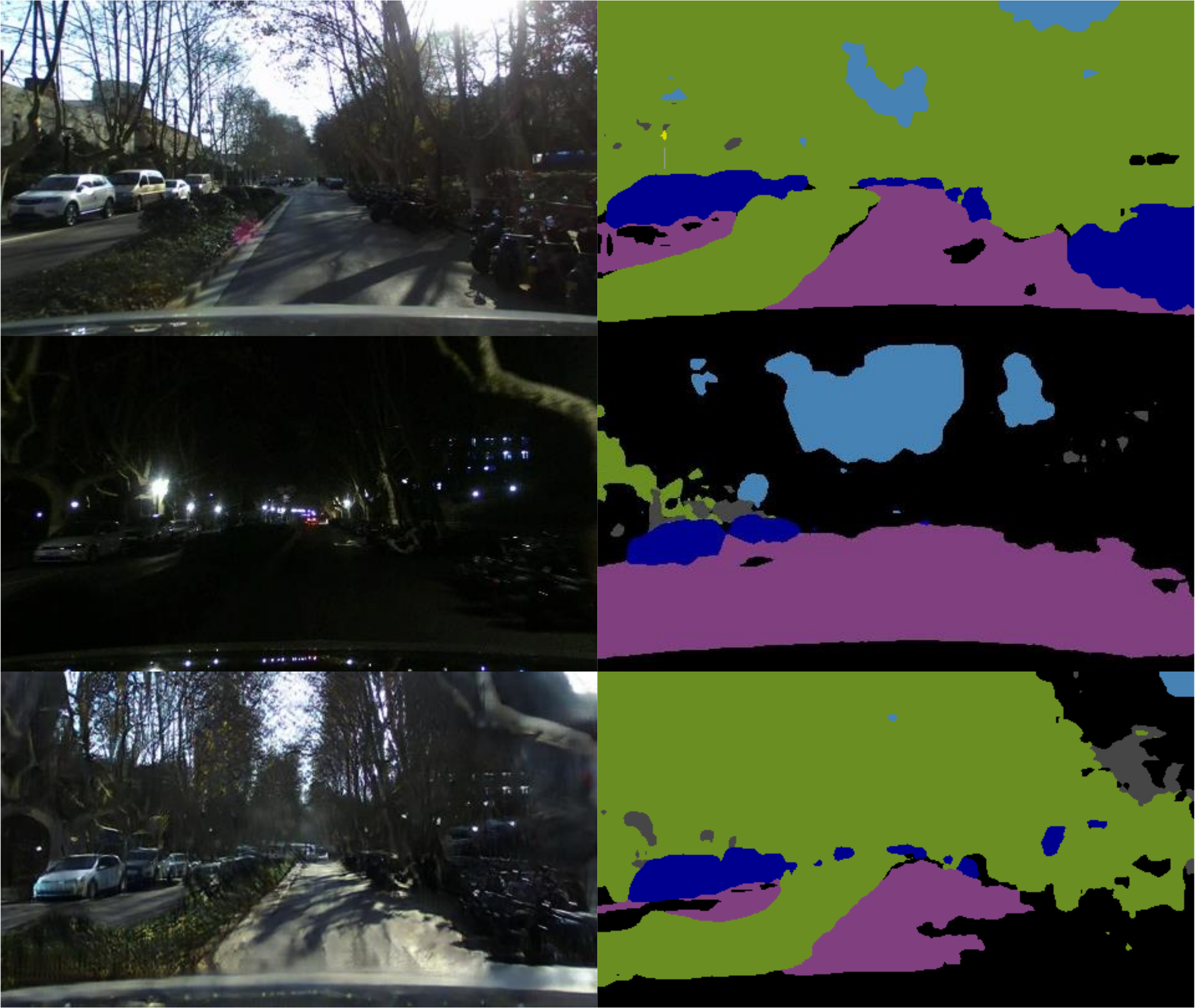}}
   \subfigure[]{
   \label{Fig.ZJU-GAN sub.2}
   \includegraphics[width=0.49\textwidth]{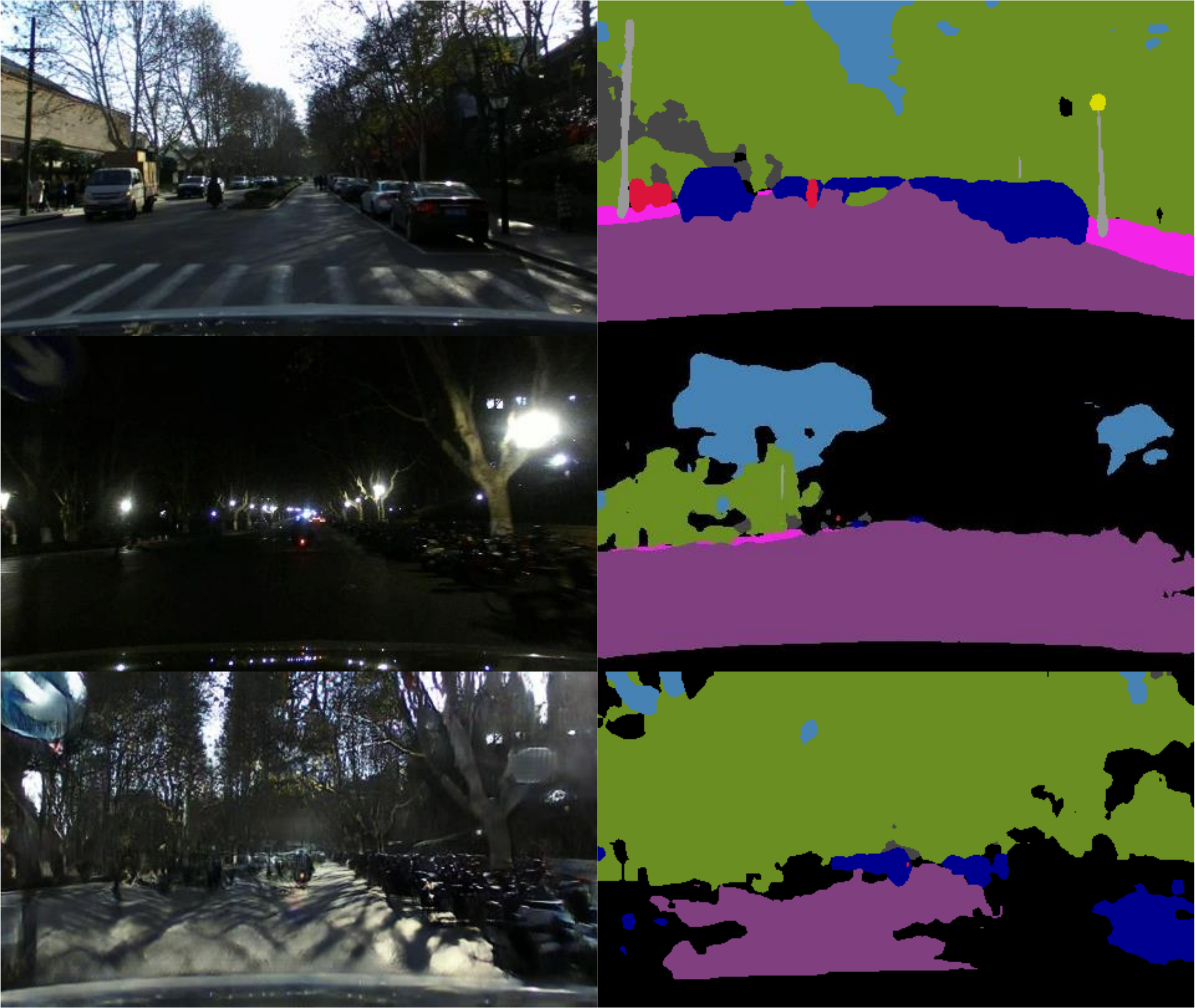}}
   
   \subfigure{
      \label{classes}
      \includegraphics[width=0.98\textwidth]{labels.png}
   }
   \caption[ZJU-GAN]
   { \label{fig:ZJU-GAN} 
   Examples from ZJU dataset (Top: day input, Mid: night input, Bottom: night-to-day converted input). In the second row, all the most of the trees and some cars are missed in labeled images due to the darkness. In the bottom row, the segmentation of cars and vegetation has improved. But model performs not so well around the corner of the images.}
\end{figure}

In the first method, we use night-to-day converter to generate synthetic daytime images, which is the comfortable domain for ERF-PSPnet. Fig.~\ref{fig:BDD-GAN} and Fig.~\ref{fig:ZJU-GAN} shows some representative results of our experiment. The first row and second row show the daytime images and nighttime images respectively, and the bottom row shows the synthetic daytime image converted form the images in second row.

Fig.~\ref{fig:BDD-GAN} shows our results in BDD Dataset. First rows of both subfigures show the stably-behaving performance of ERF-PSPNet in daytime driving images. Nearly all the classes are labeled precisely. In the second row, subfigure (a) is worse illuminated than subfigure (b). Our daytime-trained model fails to detect the sky in both images. Cars in the left are ignored by the model in subfigure (a) and the whole gas station is missed in subfigure (b). In the bottom row, the model recognizes the sky, more cars and the whole gas station in the synthetic images successfully. In this way, GANs help the model improve performance at night. However, GANs also bring some problems: textures of objects like roads and cars in synthetic images are different from the ones in the real world, causing confusion to the model. As we can see in the bottom row of subfigure (a), the left part of the road is not labeled completely.

Fig.~\ref{fig:ZJU-GAN} shows our results in ZJU dataset. Because our semantic segmentation model is trained on BDD dataset, in which the streetscape significantly varies from ZJU datset, semantic segmentation outputs in the first row are not as ideal as ones in BDD dataset due to the geographical location-related domain gap between BDD (Berkeley) and ZJU (Hangzhou). In the second row we can find that all of the bushes and motorcycles beside the road are recognized as road, which is extremely dangerous for autonomous driving. In the bottom row, synthetic daytime images perform much better than nighttime images in the second row in general (night-to-day converter helps the model to correctly label the trees and bushes, which are very dark at night), but the lost details of the synthetic images make part of the road and the traffic sign missed in the labeled images.

\begin{figure}[ht]
   \centering
   \subfigure[Input]{
   \begin{minipage}[b]{0.235\linewidth}
   \includegraphics[width=1\linewidth]{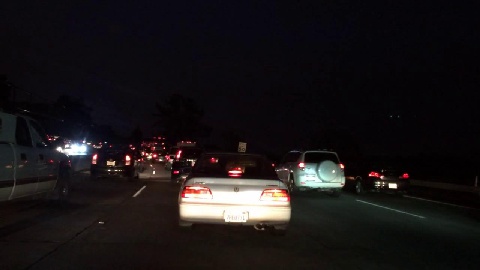}\vspace{0pt}  
   \includegraphics[width=1\linewidth]{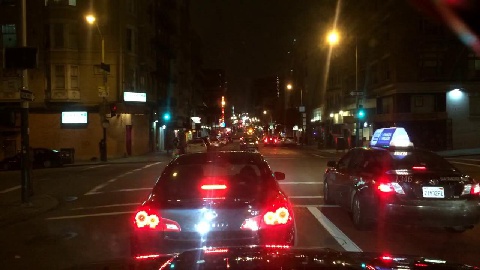}\vspace{2pt}
   \includegraphics[width=1\linewidth]{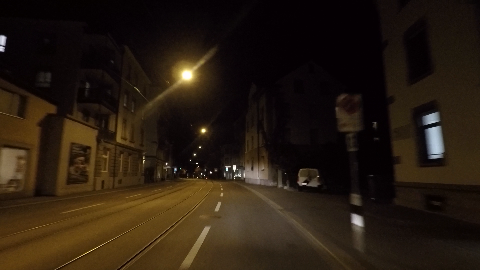}\vspace{0pt}
   \includegraphics[width=1\linewidth]{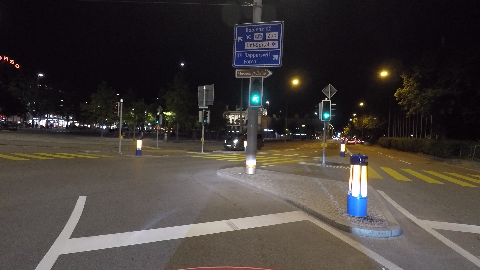}
   \end{minipage}}\hspace{-6pt}  
   \subfigure[Ground Truth]{
   \begin{minipage}[b]{0.235\linewidth}
   \includegraphics[width=1\linewidth]{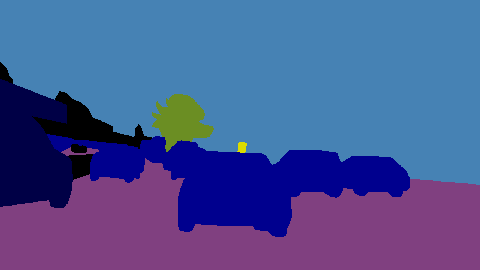}\vspace{0pt}
   \includegraphics[width=1\linewidth]{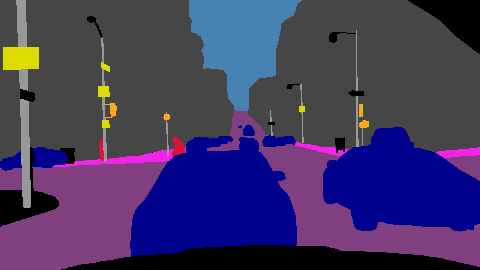}\vspace{2pt}
   \includegraphics[width=1\linewidth]{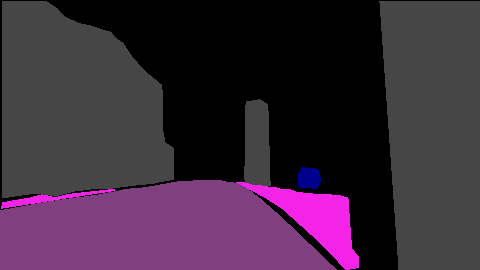}\vspace{0pt}
   \includegraphics[width=1\linewidth]{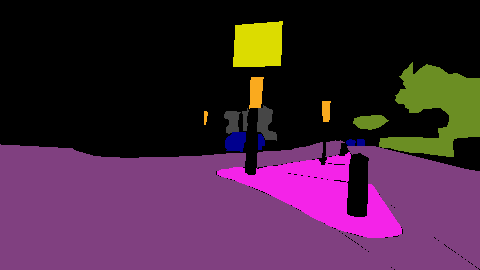}
   \end{minipage}}\hspace{-6pt}
   \subfigure[Ordinary ERF-PSPNet]{
   \begin{minipage}[b]{0.235\linewidth}
   \includegraphics[width=1\linewidth]{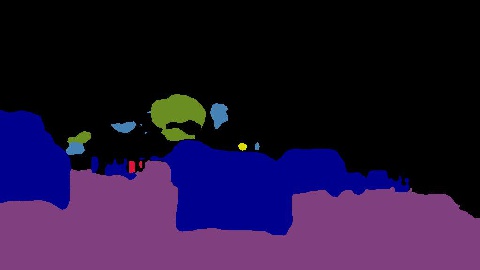}\vspace{0pt}
   \includegraphics[width=1\linewidth]{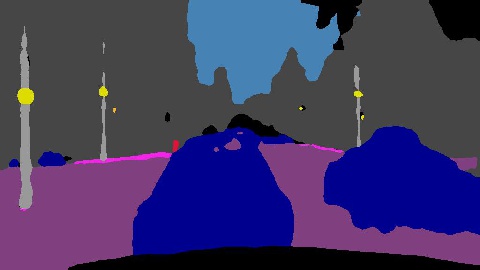}\vspace{2pt}
   \includegraphics[width=1\linewidth]{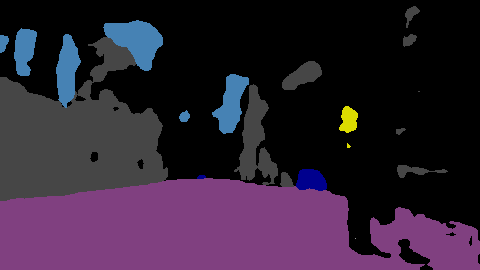}\vspace{0pt}
   \includegraphics[width=1\linewidth]{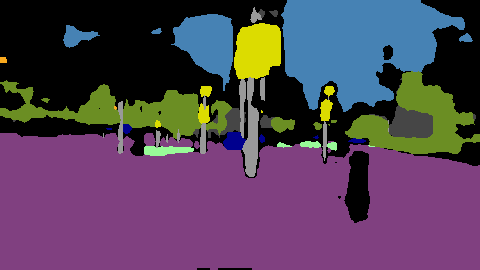}
   \end{minipage}}\hspace{-6pt}
   \subfigure[Our Method]{
   \begin{minipage}[b]{0.235\linewidth}
   \includegraphics[width=1\linewidth]{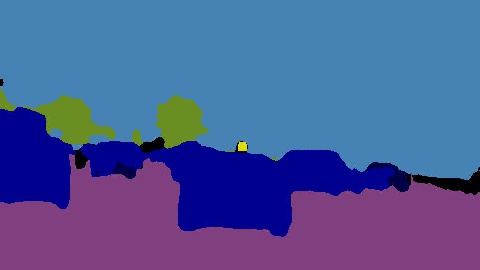}\vspace{0pt}
   \includegraphics[width=1\linewidth]{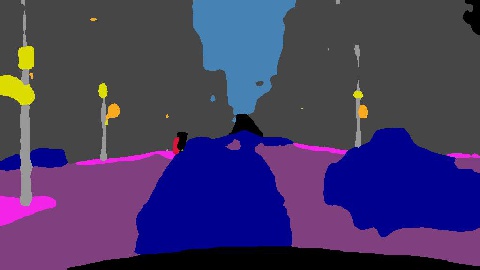}\vspace{2pt}
   \includegraphics[width=1\linewidth]{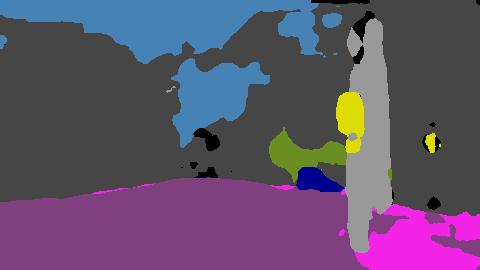}\vspace{0pt}
   \includegraphics[width=1\linewidth]{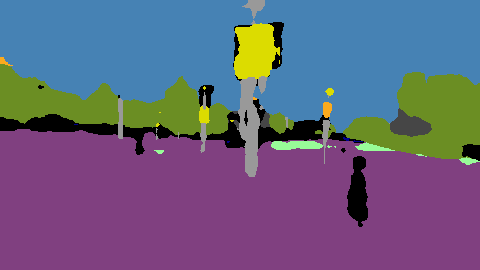}
   \end{minipage}}
   \caption[night-trained]
   { \label{fig:night-trained}
   Examples from BDD Dataset and Nighttime Driving Dataset (Top two rows: BDD Dataset, Bottom two rows: Nighttime Driving Dataset). In general, our method (Converting 2,000 images to synthtic nighttime images in training) performs better than ordinary ERF-PSPNet(trained in original BDD10K training set). All the classes especially sky are recognized better}
\end{figure}

In the second method, different ratio of images with annotated labels in training set are converted to nighttime images to improve the robustness of semantic segmentation model in face of nighttime images. Fig.~\ref{fig:night-trained} shows performance of the model on the validation set of the BDD dataset and the testing set of the Nighttime Driving Dataset. In this example, our method relies on training ERF-PSPNet with 2,000 synthetic nighttime images and 5,000 original images in the training set. Different ratios of synthetic nighttime images in training set will be discussed in the next subsection. As shown in Fig.~\ref{fig:night-trained}, the traffic signs and sky are significantly better labeled with our method. ERF-PSPNet yielded with the original training set performs decently for nighttime images with good illuminance (bottom row), but much worse under bad illuminance conditions. However, our method shows enhanced robustness in all conditions.

\subsection{Quantitative Results and Discussion}

Table.~\ref{tab:results} shows the quantitative results of our two methods and contrast method. In the table, the first three rows are results from contrast method: training ERF-PSPNet with BDD10K dataset and testing with 968 daytime images and 32 nighttime images in the BDD10K test set. The contrast method represents the baseline for our proposed two methods.

\begin{table}[ht]
   \caption{Results of our two methods in BDD val set and Nighttime Driving test set.} 
   \label{tab:results}
   \begin{center}       
   \begin{tabular}{|l|l||l|l|}
   \hline
   \rule[-1ex]{0pt}{3.5ex}  \textbf{Train/Method} & \textbf{Test} & \textbf{Mean IoU} & \textbf{Mean Acc} \\
   \hline
   \hline
   \rule[-1ex]{0pt}{3.5ex}  BDD dataset & Day in BDD test set & 52.10\% & 75.52\%  \\
   \hline
   \rule[-1ex]{0pt}{3.5ex}  BDD dataset & Night in BDD test set & 32.72\% & 75.67\%  \\
   \hline
   \rule[-1ex]{0pt}{3.5ex}  BDD dataset &  Nighttime Driving test set & 36.73\% & 72.38\%  \\
   \hline
   \hline
   \rule[-1ex]{0pt}{3.5ex}  BDD dataset & Night2day in BDD test set & 29.94\% & 56.87\%  \\
   \hline
   \rule[-1ex]{0pt}{3.5ex}  BDD dataset & Night2day in Nighttime Driving test set & 32.74\% & 66.46\%  \\
   \hline
   \hline
   \rule[-1ex]{0pt}{3.5ex}  Day2night BDD dataset & Day in BDD test set & \textbf{53.03\%} & 75.96\%  \\
   \hline
   \rule[-1ex]{0pt}{3.5ex}  Day2night BDD dataset & Night in BDD test set & \textbf{43.14\%} & 68.93\%  \\
   \hline
   \rule[-1ex]{0pt}{3.5ex}  DarkModelAdaptation~\cite{dai2018dark} & Nighttime Driving test set & 41.60\% & NA  \\
   \hline
   \rule[-1ex]{0pt}{3.5ex}  Day2night BDD dataset & Nighttime Driving test set & \textbf{45.09\%} & 72.82\%  \\
   \hline
   \end{tabular}
   \end{center}
   \end{table}

The results of first method (converting nighttime images on-the-fly to synthetic daytime images during inference), are shown in the forth and fifth rows. As we can see, the results are below the baseline. In general, this method performs worse than no methods applied, but in some classes such as sky, car and trucks, accuracies improve remarkably. The main cause is that the textures of synthetic images from GANs are not as detailed as those of natural images. Semantic segmentation model is trained on natural daytime images, causing low accuracy in labeling synthetic daytime images from the test sets. Another possible reason is that nighttime images in BDD dataset and Nighttime Driving dataset are illuminated decently, so our method may not be much better than the baseline.

The results of second method: converting parts of the daytime images in training set to nighttime images on the stage of training, are shown in last four rows, together with a baseline method DarkModelAdaptation proposed by D.~Dai et al.\cite{dai2018dark} validated on their dataset. Mean IoU increases remarkably for the nighttime images in the BDD testing set and Nighttime Driving test set, and keeps the same level of accuracy as the baseline for daytime images. Compared to the method proposed by D.~Dai et al.\cite{dai2018dark}, our method rises nearly 4\% on the same Nighttime Driving test set, even though our ERF-PSPNet is much smaller than RefineNet\cite{lin2017refinenet} adopted by them. In general, our method dramatically improves the performance of ERF-PSPNet in face of nighttime images.

\begin{figure}[ht]
   \centering
   \includegraphics[width = 0.7\textwidth]{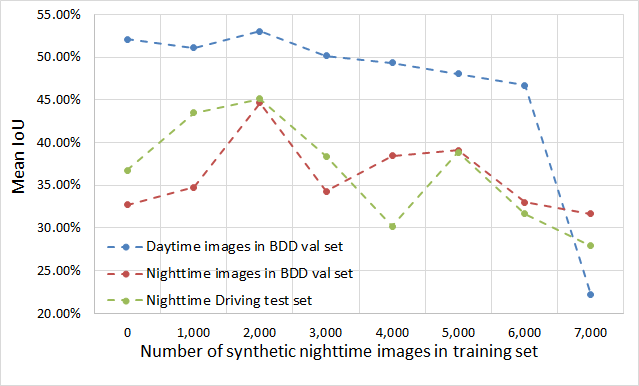}
   \caption{\label{fig:diagram}
   Number of synthetic nighttime images in training set - Mean IoU curve. IoU value peak appears at 2,000 synthetic nighttime images in the train set.}
\end{figure}

We perform an important experiment to explore how the ratio of synthetic nighttime images in the training set influences the result. Fig.\ref{fig:diagram} shows the mean IoU of our day-to-night method with respect to different ratios of synthetic nighttime images in the training dataset. When varying the number of synthetic nighttime images in the training set, we find that in the range of 0 - 2,000, as the ratio of synthetic nighttime images gets higher, the model performs increasingly better for nighttime images. The model learns well the illumination of scene elements in synthetic nighttime images and textures of real objects in daytime images. But as the synthetic nighttime images gets more, IoU gets down on the contrary. Additionally, at 5,000, the curve reaches another peak. The reason may be that 5,000 is a symmetrical number to 2,000 (7,000 in total), and the model learns the texture from daytime images and the illumination from synthetic nighttime images in a complementary way, but the daytime performance has already degraded to a low level. When all images in train set are converted to nighttime images, the IoU gets even lower than 30\% for the same reason as the low IoU of the first method: the textures in synthetic images are different from that in real images. In the end, it turns out that the sweet spot is to use 2,000 synthetic nighttime images and 5,000 real daytime images as training set, as this ratio reaches the best accuracy for nighttime semantic segmentation, while daytime semantic segmentation also remains robust.

\section{CONCLUSIONS}

In this paper, we investigate the problem of semantic image segmentation of nighttime scenes. To improve the performance, two methods are proposed by training CycleGAN as a two-way day-night converter. In the first method, nighttime images are converted to daytime domain on-the-fly during inference as a pre-processing step. In the second method, a critical part of images of BDD training set are converted to synthetic nighttime images, improving the robustness of the segmentation model in the training process.

To validate our methods, three datasets are leveraged to obtain qualitative and quantitative results. Our comprehensive set of experiments indicates the path to follow, and the sweet spot to determine the training strategy, in order to attain the best robustness across the day and night. Overall, these results demonstrate that our methods improve the model performance observably, making state-of-the-art efficient networks like ERF-PSPNet work robustly at night.

\bibliography{main} 
\bibliographystyle{spiebib} 

\end{document}